# Modality-bridge Transfer Learning for Medical Image Classification

Hak Gu Kim[†]
School of Electrical Engineering
KAIST
Republic of Korea

Yeoreum Choi[†]
School of Electrical Engineering
KAIST
Republic of Korea

Yong Man Ro[*]
School of Electrical Engineering
KAIST
Republic of Korea

*Abstract*— **This paper presents a new approach of transfer learning-based medical image classification to mitigate insufficient labeled data problem in medical domain. Instead of direct transfer learning from source to small number of labeled target data, we propose a modality-bridge transfer learning which employs the bridge database in the same medical imaging acquisition modality as target database. By learning the projection function from source to bridge and from bridge to target, the domain difference between source (e.g., natural images) and target (e.g., X-ray images) can be mitigated. Experimental results show that the proposed method can achieve a high classification performance even for a small number of labeled target medical images, compared to various transfer learning approaches.**

## I. Introduction

Medical imaging is one of the most important diagnostic tools to visually represent the anatomical structures of the human body [1]. Over the past several decades, various types of medical imaging technologies including X-ray radiography, magnetic resonance imaging (MRI), computerized tomography (CT), and ultrasonic medical imaging have appeared and matured [2-5].

In the medical imaging, machine learning techniques play a key role by helping to solve the diagnostic and prognostic problems in a variety of medical imaging fields. Many researchers have proposed machine learning-based methods for clinical parameter analysis, medical knowledge extraction, disease progression prediction, etc. The authors of [6] proposed lung segmentation in the chest X-ray images using a region-based active contour. In [7], a brain region detection related to Alzheimer's disease was proposed from brain MRI based on support vector machine (SVM). In [8], a multivariable logistic regression-based tool was proposed for screening malignant lung nodules from the low-dose CT. More recently, with the success of deep learning in computer vision application, the deep learning frameworks have been applied to the medical imaging [9-11]. However, there are limitations in adopting machine learning in the medical imaging due to unique characteristics of medical image database, e.g., incompleteness by missing parameters and the lack of publicly sufficient labeled database. In particular, a small number of labeled databases are one of the main factors that make it difficult to well train the classification model due to over-fitting problems. Although some deep convolutional neural network (CNN)-based methods achieved impressive performances in medical image domain, it is still hard to fully train deep network with a limited number of labeled datasets [10, 11].

To overcome the aforementioned limitation, many transfer learning-based methods have been proposed in medical imaging applications. The aim of the transfer learning is to learn the prediction function in the target domain using the knowledge learned by a large number of labeled data set (e.g., ImageNet [12]) in source domain. In various computer vision fields, it is well known that the transfer learning contributes to the improvement of learning of sparse labeled or a small number of dataset in the target domain [13-18]. For the transfer learning in medical imaging, however, the input image characteristics are very different between the training data (a large number of natural images) and test data (a small number of medical images). Due to the extremely different domains having different and unrelated classes, the transferred function learned from the source database (training set) can be biased when directly applied to the target database (training set) [19]. As a result, the features extracted from the biased function are unlikely to be desirable for target domain, which is medical image domain.

The aim of this paper is to propose a novel approach of transfer learning-based classification with a small number of medical images considering additional database of the same acquisition modality of target data (we call it bridge database). Conventional transfer learning could cause degraded classification performances in medical image domain due to the significant distribution mismatches between source (natural image) and target (medical image) domains. To mitigate the distribution mismatches between source and target domains, we devise a new transfer learning through the bridge database (we call it modality-bridge transfer learning). The bridge database refers to a medical image set that has different purposes but is obtained from the same medical acquisition modality. For example, a large-scale chest X-ray image set can be used as bridge database as a representative of X-ray acquisition modality for cyst classification. By transferring the learned projection functions from natural images (source) to

---

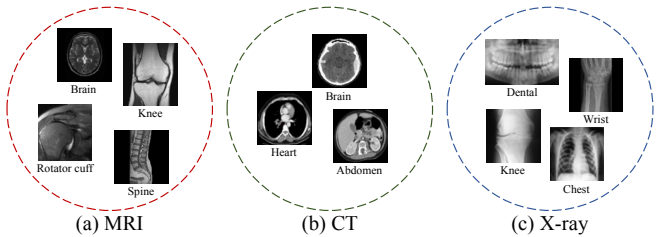

Figure 1. Examples of three different types of medical image acquisition modalities.

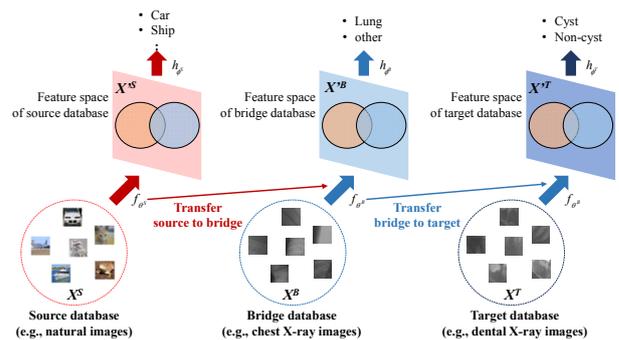

Figure. 2. Overall framework of the modality-bridge transfer learning for classification. The proposed method transfers the knowledge from source to bridge and from bridge to target to reduce the domain difference between source and target databases. The transferred knowledge of source database and the projection function learned with the bridge database can represent discriminative features in the target domain because bridge and target databases are obtained from the same medical imaging modality. With the knowledge of bridge database, the classifier of the target domain can be well trained.

chest X-ray images (bridge) and from chest X-ray images to dental X-ray images (target), the distribution mismatch (not only different domain but also small number of labeled target data) between natural image domain and dental X-ray image domain can be reduced during the modality-bridge transfer learning.

The proposed method consists of consecutive three main parts as follows: 1) learning the projection function in the source domain projecting from source image space to its feature space, 2) learning the nonlinear mapping from feature space of the source image to feature space of the bridge database based on the projection function learned by source database, and 3) learning the classifier based on the transferred projection function learned by bridge database.

To evaluate the effectiveness of the proposed method using new bridge transfer learning, extensive experiments have been conducted in various types of medical image acquisition modalities such as X-ray, MRI, and CT. Experimental results show that the classification results of the proposed method are much better than those of other approaches over a small number of medical image database.

The rest of this paper is organized as follows. Section II describes the proposed transfer learning and its learning process. In Section III, comprehensive experimental results are shown to verify the usefulness of the proposed modality-bridge transfer learning. Finally, conclusions are drawn in Section IV.

## II. PROPOSED METHOD

### A. Medical Image Configuration for the Proposed Transfer Learning

Fig. 1 shows examples of various medical images from three different medical image acquisition modalities which are MRI, CT, and X-ray. As shown in the Fig. 1, despite different human body parts, the images in the same medical image acquisition modalities are likely to have similar characteristics. For classification task in the sparse labeled or small-scale target medical images, a sufficient number of images in the same medical image acquisition modalities are obtained from multi-site as a bridge database. The bridge database could not increase target data size but lead to domain adaptation between source database (natural images) and target database (medical images).

In the proposed transfer learning with small target image, the database is composed of large-scale natural images as source database, a small number of medical images for target database (e.g., dental X-ray images with cysts), and a sufficient number of medical images of bridge database. The bridge database is obtained from same medical image acquisition modalities as the target database from multi-site (e.g., chest X-ray images).

### B. Overall Framework of the Modality-bridge Transfer Learning for Medical Image Classification

Fig. 2 shows the overall framework of the proposed modality-bridge transfer learning for medical classification using domain adaptation. To extract images characteristics such as edge and texture, we learn the projection function mapping source image space to source feature space through source database which consists of a large number of natural images. Then, the knowledge learned by natural images is transferred to the bridge database (i.e., medical images from the same medical imaging modality as the target) in order to learn the characteristics of medical images (e.g., bone and tissues in X-ray images). Finally, based on the learned characteristics of the natural and bridge medical images, the classifier is learned for the target database. Detailed descriptions on the proposed classification are given in the following subsections. Note that the proposed modality-bridge transfer learning is established for the classification task in the sparse labeled or small-scale target medical images.

### C. Domain Adaptation using Modality-bridge Transfer Learning

*1) Learning for the characteristics of the natural images using source database:* Let $x_i^S$ and $x_i'^S$ denote $i$-th source image and its feature vector, respectively. $y_i^S$ indicates the class label corresponding to $x_i^S$. Let $f_{\theta^S}$ and $h_{\phi^S}$ denote projection function with parameter $\theta^S$ and classification function with parameter $\phi^S$ in the source domain, respectively. To learn the characteristics of the natural images such as color, edge, and texture, we train the projection function mapping natural image to its feature vector through a large number of natural images. With the learned features, the classifier is

learned to predict their classes. Given input $x_i^S$, the probability of its class $y_i^S$ can be written as

$$p(y = y_i^S | x_i^S; \theta^S, \phi^S) = \frac{e^{h_{\phi^S}(f_{\theta^S}(x_i^S))}}{\sum_{j=1}^{N^S} e^{h_{\phi^S}(f_{\theta^S}(x_j^S))}}, \quad (1)$$

where $N^S$ indicates the number of source database.

With the probability term of Eq. (1), by minimizing the cross-entropy loss, the trained projection and classification functions for the classification of the source database (natural images) can be obtained. However, it does not reflect the characteristics of target database (e.g., dental X-ray images) due to the domain difference.

*2) Learning for the characteristics of the medical images using bridge database:* To learn the characteristics of the target domain (medical image), the knowledge of the source database is supposed to be transferred. As mentioned earlier, in medical imaging domain, it is difficult to collect a large number of labeled images because of the patient privacy protection and a high cost for reliable labeling. For this reason, direct transfer learning from source to target which has small number of labeled image could cause over-fitting problem or failure of the convergence of training.

In this paper, we employ the bridge database. As abovementioned in Section II.A, the bridge database consists of sufficient number of medical images from the same medical imaging modality collected from multi-center. The domain of the bridge database (e.g., chest X-ray images) is the same as the target domain (e.g., dental X-ray images). By transferring the projection function learned in the source domain to the bridge database domain, over-fitting problem can be reduced while the characteristics of the medical images are learned.

*3) Learning for the characteristics of the specific medical images using target database:* Our goal is to predict true class in the target medical image domain even for insufficiently labeled condition. In the proposed method, the target images are projected onto the target feature space through the transferred projection function of the bridge database, $f_{\theta^B}$. The discriminative features of target database can be obtained by $f_{\theta^B}$ because the bridge database is obtained from the same medical imaging modality. Obviously, there is difference between the bridge and target image domain. To mitigate this problem, only classification function in the target domain is learned because the amount of the target database is not large enough to learn the projection function. Based on the projection function of the bridge database, the classification function for target database can be learned by minimizing the cross-entropy loss function, which can be written as

$$\phi^{T*} = \arg\min_{\phi^T} \left[ -\sum_{i=1}^{N^T} y_i^T \log p(x_i^T; \theta^B, \phi^T) \right]$$
$$= \arg\min_{\phi^T} \left[ -\sum_{i}^{N^T} y_i^T \log \left( \frac{e^{h_{\phi^T}(f_{\theta^B} x_i^T)}}{\sum_{j}^{N^T} e^{h_{\phi^T}(f_{\theta^B} x_j^T)}} \right) \right], \quad (2)$$

where $x_i^T$ and $y_i^T$ represent *i*-th target image and its class, respectively. $\theta^B$ and $\phi^T$ are parameters of the projection function of the bridge database and classification function of the target database, respectively. $N^T$ is the number of target database.

Using the Eq. (2), by finding the optimal parameters, $\phi^{T*}$, of $h_{\theta^T}$ mapping the features projected by $f_{\theta^B}$ to the target domain, the classification performance in the target domain can be improved while mitigating the over-fitting problems.

III. EXPERIMENTS AND RESULTS

To verify the effectiveness of the proposed method, we performed extensive experiments in three different medical images acquisition modalities that were X-ray, MRI, and CT. In our experiments, ImageNet Large Scale Visual Recognition Challenge 2012 (ILSVRC 2012) database [12] was used as the source database. Detailed explanations of other bridge and target databases are described in the following subsections.

A. Experimental databases

*1) X-ray image acquisition modalities:* In the X-ray image modality, the target database consisted of 120 dental panoramic X-ray images for cyst classification. Two kinds of image patches, cyst and non-cyst patches, were labeled. We used 963 patches including 539 cyst patches and 424 non-cyst patches. For the bridge database, we used JSRT digital image database [20] including 247 chest X-ray images. We divided the entire images into small patches belonging to the lung part or other parts. So we used 13,119 patches containing 6,223 lung patches and 6,896 other body patches.

*2) MRI acquisition modalities:* For target database on MRI image modality, we made use of OASIS database [21] for Alzheimer's disease (AD) classification. It contained 761 brain MRI images including AD and normal cases. There were 326 AD cases (very mild, mild, moderate, and severe dementia) and 435 normal cases. In experiment to verify the proposed method, we used 1,500 patches including 720 AD and 780 normal patches. We used NCI-ISBI 2013 Challenge database [22] for the bridge database on MRI domain. There were 1,258 prostate MRI images with their segmentation maps. From this information, we constructed 18,746 patches including 7,308 prostate patches and 11,438 other part patches.

*3) CT image acquisition modalities:* In the case of CT image modality, we utilized thoraco-abdominal lymph node (LN) database [23] as the target database. There were 388 mediastinal LNs and 595 abdominal LNs from 90 and 86 patient CT scans, respectively. In the experiment, we limitedly used 1,264 patches including 448 mediastinal LNs patches and 816 abdominal LNs patches for classification of mediastinal LNs. For the bridge database on CT domain, we selected SPIE-AAPM-NCI Lung Nodule Classification Challenge database [24] including 22,489 lung CT images. It consisted of 82 malignant cases and 84 benign cases. We obtained 17,928 patches including 8,856 malignant patches and 9,072 benign patches by augmentation.

TABLE I. CLASSIFICATION PERFORMANCES WITH SMALL-SCALE TARGET DATABASES

|  | Target database modality | | |
|---|---|---|---|
|  | X-ray | MRI | CT |
| Direct transfer learning using VGG16 | 81.5% | 44.7% | 85.8% |
| Modality-bridge transfer learning with bridge data of the same acquisition modality | 90.1% | 71.4% | 91.4% |
| Transfer learning with bridge data of the different acquisition modality | 66.4% | 55.3% | 80.4% |

*B. Projection and classification function*

In our experiment, VGG16 network [25] was employed as the projection and classification function, which is trained by ImageNet dataset. There were a total of 16 layers, which are 13 convolutional layers, 2 fully-connected layers, and a softmax layer. The 13 convolutional layers and 2 fully connected layers were considered as the projection function.

*C. Experimental results*

*1) Experimental setup:* Our experiments were conducted on a PC with Intel Core i7-4770 @ 3.40GHz, 32GB RAM, and NVIDIA GTX 1080.

*2) Performance comparisons:* To evaluate the performance of the proposed method in the sparse labeled or small-scale target medical images, 963, 1,500, and 1,264 labeled image patches from X-ray, MRI, and CT were used as the target databases. For performance comparisons, we performed a direct transfer learning (source to target) and the modality-bridge transfer learning with bridge database of the same medical acquisition modality. For the direct transfer learning, the parameters of the VGG16 model trained by ImageNet were directly applied to the target database. Then, softmax layer for classification was re-trained to avoid over-fitting. For the modality-bridge transfer learning, medical images gathered from the same medical imaging acquisition modality as the bridge database (see Section III.A).

In Table I, the first and the second rows show the classification accuracy of the direct transfer learning and the proposed modality-bridge transfer learning in three medical image acquisition modalities. As shown in Table I, the modality-bridge transfer learning achieved significantly higher classification performances in three different medical image acquisition modalities than those of the direct transfer learning. In addition, we investigated the performance when the image modality of the bridge database was different from that of the target database. In these experiments, the prostate MRI, the chest X-ray, and the prostate MRI were used as bridge database for X-ray, MRI, and CT, respectively. The third row of Table I shows the performances of the transfer learning with the bridge database in the different acquisition modality. As shown in Table I, the classification accuracy is decreasing. This result indicated that the bridge database from the different medical imaging acquisition modality did not mitigate the domain differences because of the unique characteristics of each medical image modality.

Furthermore, we performed additional experiments of a classification task with sufficient number of target database applicable to direct transfer learning. We investigated the performance when the amount of target database was large enough to train the network through the direct transfer learning. In these experiments, for the target database in X-ray, we used 14,123 patches including 6,517 cyst patches and 7,606 non-cyst patches. For the target database in MRI, we used 15,220 patches including 6,520 AD patches and 8,700 normal patches. For the target database in CT, we used 15,728 patches including 6,208 mediastinal LN patches and 9,520 abdominal LN patches. The classification accuracies were 92.2% in X-ray, 73.6% in MRI, and 93.3% in CT. These results demonstrated that the proposed modality-bridge transfer learning could achieve high classification performances as much as using a large amount of target database, despite using a small-scale of the target medical database.

IV. CONCLUSIONS

In this paper, a novel approach of transfer learning for classification in medical image domain considering the bridge database was presented. By learning the projection and classification functions from source to bridge and bridge to target database, the domain differences between source and target databases could be mitigated. Experimental results demonstrated that the proposed method provided high classification performances even with a small number of target databases.